\def\year{2019}\relax
\documentclass[letterpaper]{article} 
\usepackage{aaai19}  
\usepackage{times}  
\usepackage{helvet}  
\usepackage{courier}  
\usepackage{url}  
\usepackage{graphicx}  
\usepackage{amsmath}
\usepackage{algorithm,algpseudocode}
\usepackage{threeparttable}
\usepackage{array}
\usepackage{color}

\DeclareMathOperator{\conv}{conv}
\DeclareMathOperator{\pool}{pool}
\DeclareMathOperator{\activation}{activation}
\DeclareMathOperator{\FAM}{FAM}
\DeclareMathOperator{\concat}{concat}

\begin{document}
%
\title{Amalgamating Knowledge towards Comprehensive Classification}

\author{ Chengchao Shen$^1$, Xinchao Wang$^2$, Jie Song$^1$, Li Sun$^1$, Mingli Song$^1$\\
    $^1$ Zhejiang University, $^2$ Stevens Institute of Technology\\
    \texttt{\{chengchaoshen,sjie,lsun,brooksong\}@zju.edu.cn}, \texttt{xinchao.wang@stevens.edu}
}
\maketitle
\begin{abstract}

With the rapid development of deep learning, 
there have been an unprecedentedly large number 
of trained deep network models available online. 
Reusing such trained models can
significantly reduce the cost of training the new
models from scratch, 
if not infeasible at all as the annotations used for the training 
original networks are often unavailable to public.
We propose in this paper to study a new model-reusing task,
which we term as \emph{knowledge amalgamation}. 
Given multiple trained teacher networks, 
each of which specializes in a different classification problem, 
the goal of knowledge amalgamation is to learn a lightweight student model capable of 
handling the comprehensive classification. 
We assume no other annotations except the outputs from the  
teacher models are available,
and thus focus on extracting and amalgamating knowledge 
from the multiple teachers.
To this end, we propose a pilot two-step strategy to tackle the knowledge amalgamation
task, by learning first the compact feature representations from teachers and then 
the network parameters in a layer-wise manner so as to build the student model.
We apply this approach to four public datasets and obtain very encouraging results:
even without any human annotation, the obtained student model is competent to handle the  
comprehensive classification task and in most cases outperforms the teachers in individual
sub-tasks.

\end{abstract}
\section{Introduction}

Recent years have witnessed the unprecedented progress of deep learning.
Many deep models, such as AlexNet~\cite{krizhevsky2012imagenet}, VGG~\cite{simonyan2014very}, GoogLeNet~\cite{szegedy2015going}, and ResNet~\cite{he2016deep}, have been proposed and applied to
almost every single computer vision task, yielding 
state-of-the-art performances. 
However, the promising results come with the costs of
the huge amount of  annotations required
and the resource-consuming training process, 
which may take up to weeks on multiple GPUs.

Yet encouragingly, many researchers have published online their trained models, 
tailored for various tasks like classification, detection, and segmentation.
Reusing these trained models, either for the primary task or novel ones, 
can significantly reduce the effort to retrain the new models from scratch, 
which is in many cases not feasible at all as the annotations 
used to train the original models may not be publicly available.

To this end, researchers have started to look at prospective approaches to reuse trained deep models.
For example, \cite{buciluǎ2006model} proposes a model compression approach that trains 
a neural network using the predictions of an ensemble of heterogeneous models trained a priori. 
\cite{hinton2015distilling} introduces the concept of Knowledge Distillation~(KD),
whose goal is to derive a compact student model that imitates 
the teacher by learning from the teacher's outputs. 
\cite{romero2015fitnets} makes one step further by learning 
a student model that is deeper and thinner than the teacher so as to improve the performance. 
These model-reusing approaches, despite their very promising results, 
focus on tackling the same task as the trained teacher models.

In this paper, we propose to investigate a new model-reusing task,
which we term as \emph{knowledge amalgamation}.
Given multiple trained teacher models, 
each of which specializes in a different classification problem, 
knowledge amalgamation aims to learn a  compact student model 
capable of handling the comprehensive classification problem.
In other words, the classification problem addressed by the student is 
the superset of those by all the teachers.
For example, say we have two teacher classifiers, 
the first one classifies sedan cars and SUVs 
while the second classifies pickups and vans. 
The student model is expected to be able to 
classify all the four types of cars simultaneously.
Note that, here we assume \emph{no human annotations} 
and only the predictions from the teacher models are available.

The proposed {knowledge amalgamation} task is, 
to our best knowledge, both novel and valuable. It is novel because, 
in contrast to prior model-reusing tasks that 
restrict the student model to handle the same problem as the teachers do, 
knowledge amalgamation learns the ``super knowledge''
covering the specialties from all the teachers.
It is valuable because, it allows reusing the trained models,
without any human annotation,
to learn a compact student model that approximates 
or even outperforms the teacher models.

We also propose a pilot strategy towards solving the knowledge amalgamation task. 
Our approach comprises two steps, feature amalgamation and parameter learning.
The feature amalgamation step first extracts features of the multiple teachers, 
obtained by feeding input samples to the teachers, 
and then compresses the stacked features into a  compact and discriminative set.
The obtained  set of features are then used as the supervision information for learning
the network parameters in a layer-wise manner in the  parameter learning step.
This strategy turns out to effective, as the learned 
compact student model, without any human-labeled annotations,
is capable of  handling the  comprehensive classification task
and achieves performances superior to those of the teachers on  individual
sub-tasks.

Our contribution is thus introducing the knowledge 
amalgamation task and a simple yet competent approach towards solving it,
as demonstrated on several datasets. 
We would like to promote, via the introduction of the knowledge amalgamation task,
that researchers should look at reusing trained models 
to novel tasks, in which way the 
annotation-, training-, and running-cost can be dramatically reduced.

\section{Related Work}

\subsubsection{Knowledge Distillation}

Hinton et al.~\cite{hinton2015distilling} proposes a teacher-student paradigm 
where a smaller student network imitates the soft prediction of the large teacher ones. 
This method introduces a temperature concept to highlight the similarities among 
categories, benefiting the learning of student network. 

Following \cite{hinton2015distilling}, FitNet~\cite{romero2015fitnets} adopts a 
deeper but thinner student network to learn the knowledge of a teacher. 
To improve the optimization of deep student network,  
not only the soft prediction but also the intermediate representation 
are taken into consideration to supervise the training of the student network.  
Specifically, the intermediate representation includes both the feature maps 
from the convolutional layers and the feature vectors from the intermediate 
fully connected layers. 

DK$^2$PNet~\cite{wang2016accelerating} introduces a dominant convolutional kernel method 
to compress convolutional layers. AT~\cite{Zagoruyko2017AT} 
exploits two types of spatial attention maps, activation-based 
and gradient-based from teacher network, to guide the learning of student network. 
NST~\cite{NST2017} regards the knowledge distillation as a distribution matching problem,
where the student network is trained to match the distribution of intermediate representation
with that of the teacher network. 

The knowledge distillation task, 
despite its solid motivation and proven significance, 
has a major goal-wise limitation.  
{It aims at learning a student model only from one teacher 
and thus expects the student to master only the  specialization  from  that teacher.} 
By contrast, the proposed knowledge amalgamation task enables 
the student to learn from multiple teacher models and amalgamates
all of their knowledge so as to handle the ``super'' task.

In addition, the work of~\cite{NST2017} demonstrates that 
when the number of classes is large, 
the variants of knowledge distillation approaches 
\cite{romero2015fitnets,wang2016accelerating,Zagoruyko2017AT,NST2017}
yield worse classification performances than the original version of~\cite{hinton2015distilling}.
Such variants thus do not fit our purpose, as we aim to amalgamate from multiple teachers
with potentially large number of classes shown in our experiments.

\subsubsection{Transfer Learning}
{
    Transfer learning is proposed to transfer knowledge from source domain to target domain so as to reduce
    the demand for labeled data on the target domain~\cite{pan2010survey}.
    It can be roughly categorized into cross-domain~\cite{long2013transfer,huang2018domain,hu2015deep,ding2018graph}
    and cross-task transfer learning~\cite{hong2016learning,cui2018large,gholami2017punda}.
    More specifically, cross-domain transfer learning aims to transfer knowledge among datasets with
    different data distributions but the same categories. 
    And cross-task transfer learning tries to alleviate the deficit of data for categories on the target task
    by transferring knowledge from other categories on the source task.
    However, knowledge amalgamation focuses on amalgamating the existing models with unlabeled data to 
    obtain a versatile neural network.

    Cross-modal transfer learning~\cite{ijcai2017-263,gupta2015cross,xu2018pad} transfers knowledge among different
    modalities to improve the performance on the target modality with the same categories, which is different from
    knowledge amalgamation. 
    FMR~\cite{YangZFJZ17} is proposed to introduce extra features into Convolutional Neural Network (CNN) to improve the performance on the original
    classification task.
    It is different from knowledge amalgamation, which amalgamates multiple teachers for the comprehensive classification
    task instead of the original one.
}


\begin{figure*}[t]
    \centering
    \includegraphics[width=\linewidth]{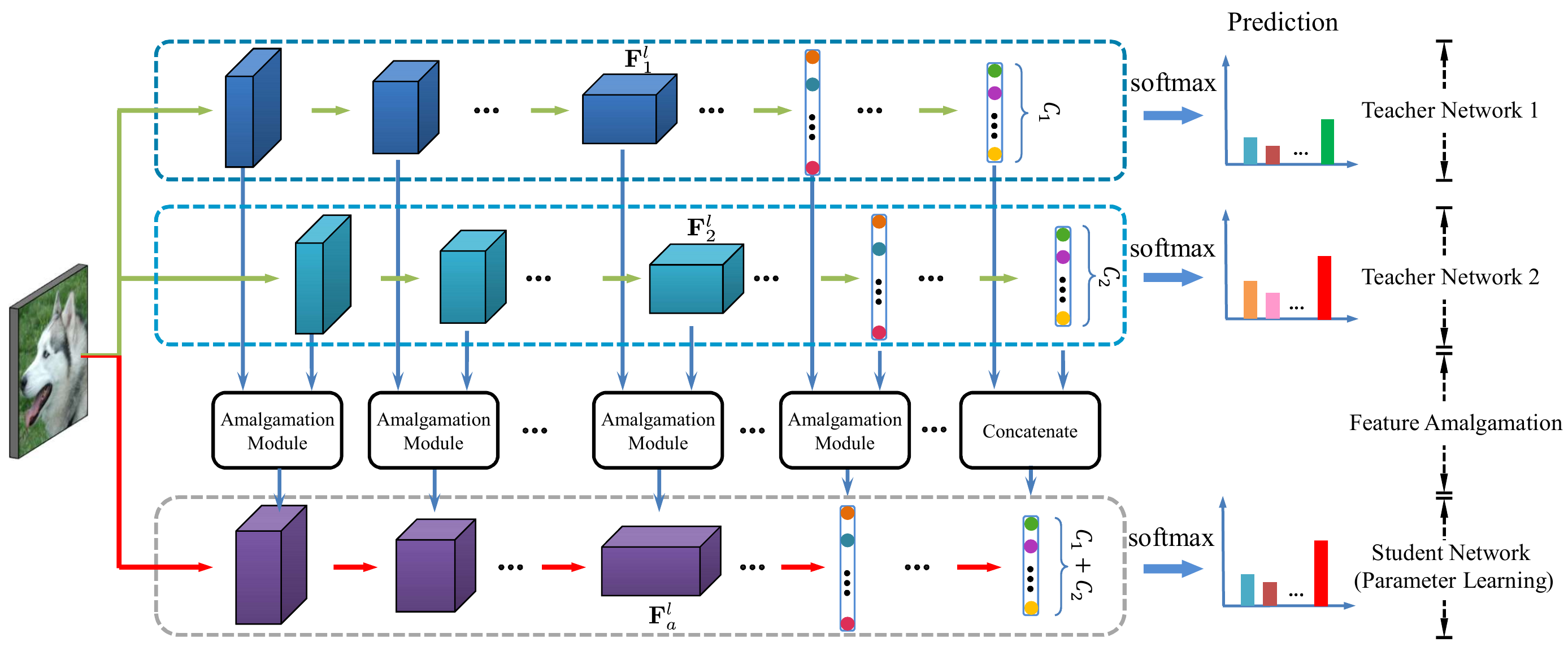}
    \vspace{-0.8cm}
    \caption{The overall workflow of the proposed approach 
    in the two-teacher case. 
    It consists of two steps, feature amalgamation and parameter learning.
    The feature amalgamation step, as depicted by the {black block}, computes the features of the 
    student model from those of the teachers.
    For example, the feature maps $F_1^l$ and $F_2^l$ from the teachers are fed into
    the  feature amalgamation module to obtain the compact feature map $F_a^l$ of the student.
    The parameter learning step,  as depicted by the 
    red arrows,   computes the network parameters of the student network, 
    given the features amalgamated from the first step.}

    \label{fig:architecture}
\end{figure*}

\section{Knowledge Amalgamation Task}

We give the definition of the knowledge amalgamation task as follows.
Assume that we are given $N$ teacher models $\{t_i\}_{i=1}^N$ trained a priori,
each of which implements a specific classification problem.
Let $\mathcal{D}_i$ denote the set of classes handled by model $t_i$. 
Without loss of generality, we assume 
$\mathcal{D}_i \neq \mathcal{D}_j, \forall i \neq j$. 
In other words, for any pair of models $t_i$ and $t_j$,
we assume they classify different sets of classes.
The goal of knowledge amalgamation is to derive a compact student model
that is able to conduct the comprehensive classification task, in other words,
to be able to simultaneously classify all the classes in 
$\mathcal{D} = \cup^{N}_{i=1} D_i$.

The student model is thus expected to be more powerful 
as it handles the ``super'' classification problem, 
and meanwhile more portable as it is smaller and more 
resource-efficient than the ensemble of the teacher models.

\section{The Proposed Method }
Towards solving the proposed
knowledge amalgamation task,  a simple yet effective pilot approach is introduced.
In what follows, we  first give an overview of the method, then detail the two steps, and finally 
show the training objective.

\subsection{Overview}

The pilot approach assumes that, for the time being, 
the teacher models share the same network architecture.  
This assumption might be arguably strong but it does hold 
in many cases especially on large-scale datasets,
where multiple models are trained on subsets of the classes.
We hope this proposed approach could serve as a baseline method 
towards solving the knowledge amalgamation, 
based on which further research could improve.
Specifically, the proposed approach follows a two-step procedure: 
{feature amalgamation} and {parameter learning}. 
In the {feature amalgamation} step, 
we derive a set of  learned features  for each layer of each teacher model, 
obtained by feeding input samples to each such teacher.
The features from the same layer across different teachers
are then concatenated and further compressed into a compact set,
which is treated as the corresponding feature map for the student.
In the parameter learning step, we treat the obtained feature sets as 
the supervision information for learning the parameters of the student network. 
This is achieved by looking at the feature sets from two consecutive layers 
and then computing the corresponding network parameters between them.

The overall process of the knowledge 
amalgamation, in the case of two teacher models, 
is shown in Figure~\ref{fig:architecture}.
The details of the feature amalgamation step and the parameter learning step 
are given as follows.


\subsection{Feature Amalgamation}
We start by discussing first the feature amalgamation from two teacher models,
and then two possible solutions for multi-teacher feature amalgamation,
followed by the score-vector amalgamation.

\subsubsection{Amalgamation from Two Teacher Models}

We first consider the case of feature amalgamation from two teacher models. 
A straightforward amalgamation approach would be to directly concatenate 
the feature sets, obtained by feeding inputs to the teacher models,
on the same layer of the two teachers. 
In this way, however, the obtained student model would be very cumbersome:
the student would be four times as large as the teachers, 
as between the two layers we will 
have twice as many the inputs and twice the outputs.

\begin{figure}[t]
    \centering
    \begin{center}
        \includegraphics[width=\linewidth]{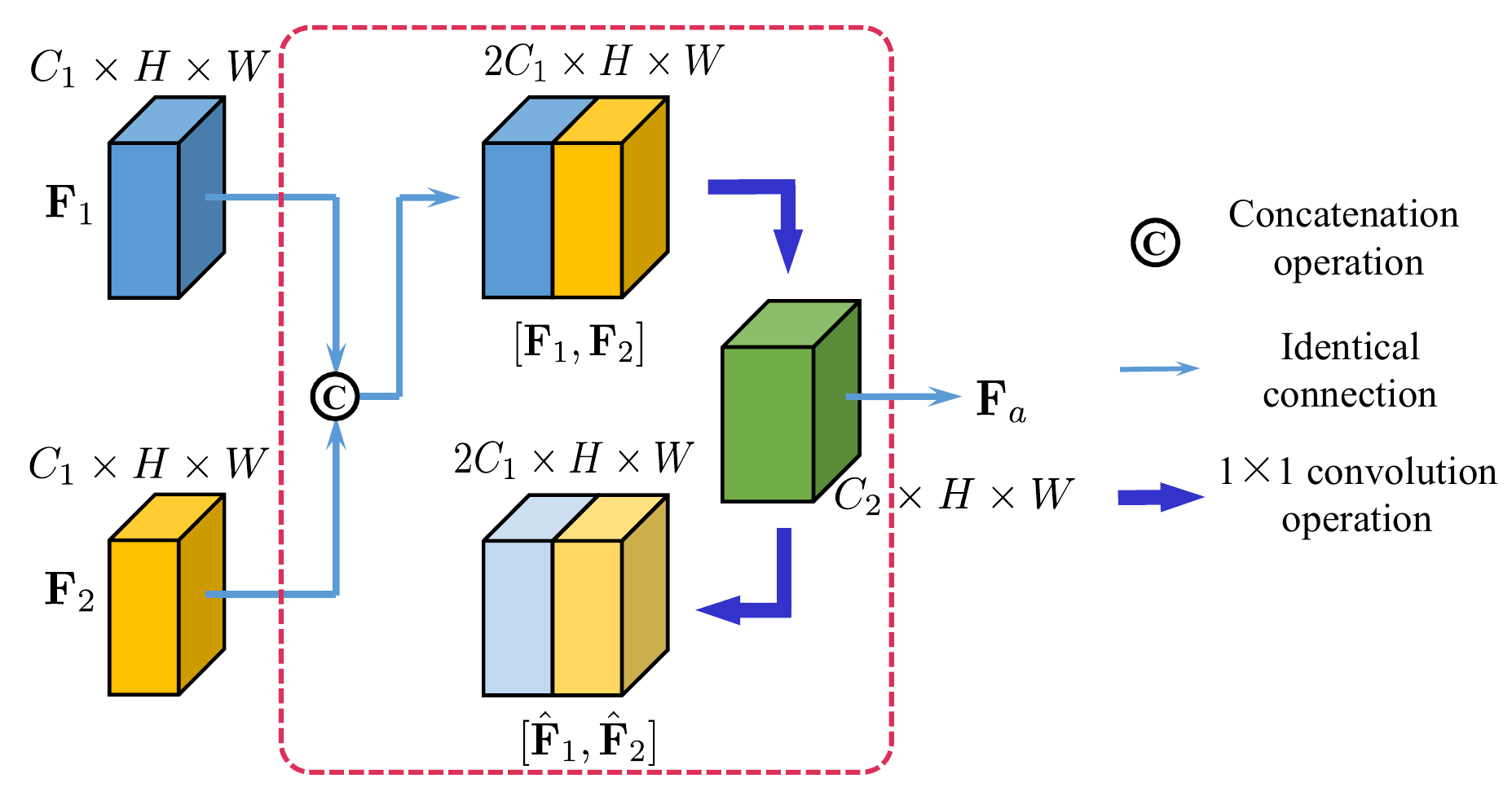}
    \end{center}
    \vspace{-0.4cm}
    \caption{Feature amalgamation from two teacher models. 
    $F_1$ and $F_2$ respectively denote the features from two teacher models,
    each of which has     $C_1$ channels.
    They are further encoded to a compact feature map $F_a$ 
    of $C_2$ channels, where $C_1 < C_2 < 2C_1$.
    The whole module is trained using an auto-encoder architecture that enforces 
    $F_a$ to preserve the information of input features.    }
    \vspace{-0.4cm}
    \label{fig:feature-amalgamation-2}
\end{figure}

Recall that the goal of amalgamation is to obtain a compact model 
that is more resource-efficient and thus handy to deploy.
To this end, we apply an auto-encoder architecture that 
compresses the concatenated features from the two teachers,
as depicted in Figure~\ref{fig:feature-amalgamation-2}.
We choose auto-encoder because it reduces the size 
of the feature maps and meanwhile preserves the 
critical information, as the compact features approximately reconstruct
the original concatenated one.

A convolution kernel of $1 \times 1$ that has demonstrated its 
success~\cite{szegedy2015going,he2016deep} in many state-of-the-art CNN architectures is adopted 
to implement the auto-encoder.
This kernel is used to reduce the channel number of feature maps
and the computation load, and meanwhile preserves the  size of the receptive field.
We write
\begin{align} \label{eq:encode}
    F_{a,c} = \sum\limits_{c'=1}^{C_{in}} w_{c, c'} \cdot F_{c'},
\end{align}
where $w_{c, c'}$ denotes the $c'$-th channel weight of the $1 \times 1$ convolution kernel, 
$c \in \{1, \dots, C_{out}\}$, $F_{c'}$ denotes the $c'$-th channel  of the input feature map $F$, 
$F_{a,c}$ denotes the $c$-th channel  of the output feature map $F_a$, 
$C_{in}$ and $C_{out}$ denote the channel numbers of input feature map $F$ 
and output feature map $F_a$, respectively.  Note that, we have 
$C_{out} < C_{in}$ due to the feature compression.

\begin{figure*}[t]
    \centering
    \vspace{-0.4cm}
    \begin{center}
        \includegraphics[width=\linewidth]{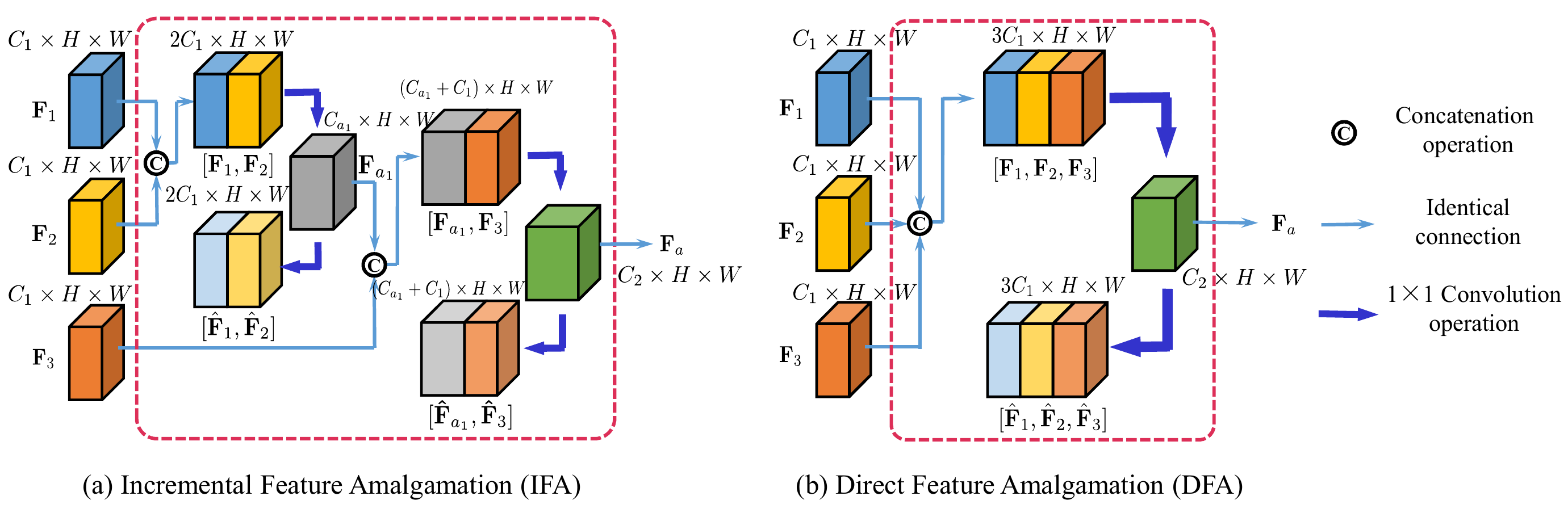}
    \end{center}
    \vspace{-0.4cm}
    \caption{Feature amalgamation from multiple teacher models. 
    (a) IFA amalgamates features progressively by each time looking at two teachers, while (b) DFA amalgamates features from multiple teachers in one shot. More specifically, IFA first amalgamates the features $F_1$ and $F_2$ to obtain features $F_{a_1}$, which is then amalgamated with $F_3$ to obtain final feature set $F_a$; DFA, on the other hand,
   simultaneously amalgamates   $F_1$, $F_2$ and $F_3$.}
    \label{fig:feature-amalgamation-3}
\end{figure*}

\subsubsection{Amalgamation from Multiple Teacher Models} 
Two ways to amalgamate features from more than two teacher models are proposed as follows.
\begin{itemize}
\item {\bf Incremental Feature Amalgamation (IFA)}: 
we conduct amalgamation in a progressive manner, by each time
amalgamating two sets of feature maps, as depicted in Figure~\ref{fig:feature-amalgamation-3} (a).

\item {\bf Direct Feature Amalgamation (DFA)}:
we directly amalgamate  feature maps from multiple teachers, as depicted in
Figure~\ref{fig:feature-amalgamation-3} (b). Similar to the two-teacher case, the feature sets are concatenated into one and then passed through an auto-encoder to obtain a compressed feature set.
\end{itemize}

Although the architecture of DFA is more intuitively straightforward, 
IFA is in fact easier to generalize as the same auto-encoder can 
be repetitively adopted and thus extended to arbitrary number of teachers,
while DFA needs to retrain the whole auto-encoder when a new teacher is added.

\subsubsection{Amalgamation of Score Vectors}
\label{sec:score_vector}

The score vector can be regarded as the response scores of 
the categories to be classified.  
For disjoint teacher models that handle non-overlapping classes, 
we directly concatenate their score vectors as the amalgamated one,
and use the amalgamated score vector as the target for the student, {as shown in Figure~\ref{fig:architecture}}.
In fact, the same strategy can also be used for teachers with overlapping classes,
in which case we treat the multiple entries of the overlapping categories
in the concatenate score vector as different classes 
during training but as the same one at test time. 
{We also test the results of preserving only 
one entry and removing the rest entries
of each overlapping category, 
which can be found in our supplementary material.}

\subsection{Parameter Learning}
In the parameter learning stage, the obtained compact
feature maps in consecutive layers are treated 
as the supervision information 
to learn the weights in between. 
Specifically, this is achieved 
by first learning the weights in a layer-wise manner and then 
fine-tuning all the layers jointly. 
To facilitate the layer-wise training,
a feature adaptation strategy is adopted.

\subsubsection{Layer-wise Parameter Learning}

Let $F_a^{l}$ and $F_a^{l-1}$ respectively denote the compact features 
of the $l$-th and $(l-1)$-th layer in the student network obtained by feature amalgamation.
In the layer-wise parameter learning step,
$F_a^{l-1}$ is fed as input and goes through a series of 
operations including pooling, activation and convolution
to approximate $F_a^{l}$. 
We write
\begin{align}\label{eq:layer_amalga}
    \hat{F}_a^l = \conv(\pool(\activation(F_a^{l-1}))),
\end{align}
where $\hat{F}_a^l$ corresponds to the estimated features in the $l$-th layer.
Since the pooling layer and activation layer have no parameters, 
$\pool(\activation(F_a^{l-1}))$ is deterministic for a given $F_a^{l-1}$.
Therefore, the goal of the layer-wise learning stage
is to obtain the weights of the convolutional layer.
This leads to a linear optimization problem, 
which is much easier to be solved than 
optimizing all the parameters of the network jointly.

\subsubsection{Feature Adaption}
A straightforward way to compute the weights of the convolutional layers 
is to solve directly the linear transformation that maps  
$\pool(\activation(F_a^{l-1}))$ to $F_a^{l}$. 
This however turns out to be sub-optimal, as $F_a^{l-1}$ 
is obtained directly from feature amalgamation and is fixed,
meaning that  $F_a^{l-1}$  is not adjustable to the non-parametric operations like pooling
and activation. 
As such, the  non-parametric layers may remove some discriminant
information from $F_a^{l-1}$, making the parameter learning troublesome.
For example, the ReLU layer will suppress all the non-positive values from the feature map $F_a^{l-1}$, 
which might be the critical information to be passed to $F_a^{l}$.

To facilitate the learning,  we introduce a Feature Adaption Module (FAM) to the 
layer-wise parameter learning stage, and transform the features into 
a form that can be well adaptive to other non-parametric layers. 
Specifically, a $1 \times 1$ convolution operation is adopted to implement FAM.
We write
\begin{align}
    \hat{F}_a^l = \conv(\pool(\activation(\FAM(F_a^{l-1})))).
\end{align}

\subsubsection{Joint Parameter Learning}
The layer-wise learning yields errors in the optimization stage, 
which accumulates layer by layer across the whole deep network. To remedy this, after the 
layer-wise learning, we look at all the parameters simultaneously and train
them end to end, in which way the convolutional layers adopt to each other better.


{
    \setlength{\textfloatsep}{2pt}
    \renewcommand{\algorithmicrequire}{\textbf{Input:}} 
    \renewcommand{\algorithmicensure}{\textbf{Output:}}
    \begin{algorithm}[!t]
      \begin{algorithmic}[1]
              \Require{N trained  teacher models  $T=\{t_i\}_{i=1}^N$,
              and unlabeled samples  $\mathcal{D} = \{x_k\}_{k=1}^K$.}
        \Ensure{The parameters of the student model $S$: $\{\Theta_l\}_{l=1}^L$}
        \For{$l=1$ to $L-1$}
            \State Obtain $\{F_i^l\}_{i=1}^N$ from $\{t_i\}_{i=1}^N$ with $\mathcal{D}$;
            \State Amalgamate feature maps  $\{F_i^l\}_{i=1}^N$ to obtain $F_a^l$;
            \State {Compute the output of $S$ from $l$-th layer: $\hat{F}_a^l$;}
            \State Compute the loss $\mathcal{L}_{\rm{PL}}^l$ according to Eq.~\ref{eq:kd};
            \State Update the parameters $\Theta_l$ using SGD;
        \EndFor

        \State Obtain $L$-th layer score vectors $\{F_i^L\}_{i=1}^N$ from $\{t_i\}_{i=1}^N$; 
        \State Obtain score vector for $S$: $F_a^L \leftarrow \concat(\{F_i^L\}_{i=1}^N)$;
        \State {Compute the output of $S$ from $L$-th layer: $\hat{F}_a^L$;}
        \State Compute the loss $\mathcal{L}_{\rm{PL}}^L$ according to Eq.~\ref{eq:kd};
        \State Jointly update the parameters $\{\Theta_l\}_{l=1}^L$ using SGD.
      \end{algorithmic}
        \caption{Knowledge Amalgamation from Multiple Teachers}
        \label{alg:A1}
    \end{algorithm}
}

\subsection{Loss Functions}
\subsubsection{Loss of Feature Amalgamation}
Recall that the target of feature amalgamation step is to remove redundant information of the concatenated
features and obtain a compact feature map that preserves the critical information of the multiple teachers.
Our objective is therefore set to be the $L_2$ loss to reconstruct the origin feature maps of $l$-th layer, as follows:
\begin{align}
    \mathcal{L}_{\rm{FA}}^l = \frac{1}{2} \Vert [\hat{F}_1^l, \dots, \hat{F}_N^l] - [F_1^l, \dots, F_N^l] \Vert^2.
\end{align}

\subsubsection{Loss of Parameter Learning}
The loss function of the parameter learning stage,
including both layer-wise learning and joint learning,
is taken to be 
\begin{align}\label{eq:kd}
    \mathcal{L}_{\rm{PL}}^l = \frac{1}{2} \Vert \hat{F}_a^l - F_a^l \Vert^2 ,
\end{align}
where $\hat{F}_a^l$ and $F_a^l$ correspond to the compact
feature maps for layer-wise learning and to the score vector
for joint learning.
The complete algorithm for knowledge amalgamation   
from multiple teacher models is summarized in Algorithm~\ref{alg:A1}, where SGD stands for Stochastic Gradient Descent.

\section{Experiments}
To evaluate the effectiveness of our proposed method, we conduct experiments on several publicly available benchmarks.
More experimental results can be found in the supplementary material.

{
    \begin{table*}
    \begin{center}
    \vspace{-0.4cm}
    \caption{Performance of {knowledge amalgamation} from two teachers on comprehensive classification task.  The best accuracy is marked in \textbf{bold} font.}
    \vspace{0.2cm}
    \label{table:two_teachers}
    \begin{threeparttable}
    \begin{tabular}{p{3.5cm}<{\centering}|p{1.3cm}<{\centering}p{1.3cm}<{\centering}|p{1.3cm}<{\centering}p{1.3cm}<{\centering}|p{1.3cm}<{\centering}p{1.3cm}<{\centering}|p{1.3cm}<{\centering}p{1.3cm}<{\centering}}
    \hline
    \multicolumn{1}{c|}{} & \multicolumn{2}{c|}{\textbf{Stanford Dogs}}    & \multicolumn{2}{c|}{\textbf{CUB-200-2011}}     & \multicolumn{2}{c|}{\textbf{FGVC-Aircraft}}  & \multicolumn{2}{c}{\textbf{Cars}}\\
    \textbf{Method} & \textbf{Params} & \textbf{Accuracy} & \textbf{Params} & \textbf{Accuracy} & \textbf{Params} & \textbf{Accuracy} & \textbf{Params} & \textbf{Accuracy}\\ 
    \hline
    {Ensemble} & $\sim$114.4M & 43.5\% & $\sim$114.8M  & 41.4\% & $\sim$114.4M & 47.1\% & $\sim$114.8M & 37.8\% \\
    \hline
    {Baseline} & $\sim$69.9M & 10.4\% & $\sim$70.2M  & 30.0\% & $\sim$69.8M & 39.9\% & $\sim$70.2M & 17.0\% \\
    \hline
    {{Layer-wise Learning}}& $\sim$69.9M & 38.4\% & $\sim$70.2M  & 31.8\% & $\sim$69.8M & 39.8\% & $\sim$70.2M & 33.6\% \\
    \hline
    {{Joint Learning}} & $\sim$69.9M & \textbf{45.3}\% & $\sim$70.2M  & \textbf{42.6}\% & $\sim$69.8M & \textbf{49.4}\% & $\sim$70.2M & \textbf{40.6}\% \\
    \hline
    \end{tabular}
    \end{threeparttable}
    \end{center}
    \vspace{-0.5cm}
    \end{table*}
}

\subsection{Experimental Setup}
\subsubsection{Dataset}
The first two datasets we adopt, CUB-200-2011~\cite{WahCUB_200_2011} and Stanford Dogs~\cite{KhoslaYaoJayadevaprakashFeiFei_FGVC2011}, 
are related to animals and the last two, FGVC-Aircraft~\cite{maji13fine-grained} and 
Cars~\cite{KrauseStarkDengFei-Fei_3DRR2013},
are related to vehicles. 
CUB-200-2011 consists of 11,788 images from 200 bird species, 
Stanford Dogs contains 12,000 images about 120 different kinds of dogs,
FGVC-Aircraft consists of 10,000 images of 100 aircraft variants, and
Cars comprises 16,185 images of 196 classes of cars. 

For each dataset, we randomly split 
their categories, which are considerably correlated, 
into parts of equal size to train two networks.
These networks are regarded as teachers to guide the learning of student network that
recognizes all categories.  
In our supplementary material, we also show the amalgamation results  
across different datasets where the categories are uncorrelated.

\subsubsection{Implementation}
The proposed method is implemented 
using PyTorch~\cite{paszke2017automatic} on a Quadro P5000 16G GPU. 
In our experiment, all the teacher models adopt the same  AlexNet architecture~\cite{krizhevsky2012imagenet},
obtained by finetuning the  ImageNet pretrained models\footnote{\url{https://download.pytorch.org/models/alexnet-owt-4df8aa71.pth}}.
The student model has a very similar network architecture as teachers. 
The only difference is that the student model has in each layer a different number of kernels, 
i.e., a different number of feature map channels.
Intuitively, the number of kernels within the student model should be 
larger than that of the teachers, as the student is more ``knowledgeable'' 
than the teachers due to its capability of handling the whole set of classes,
and meanwhile be smaller than the sum of all teachers, as it is assumed that
the features from teachers share redundancies. 
Please refer to the supplementary material for the detailed 
configuration of the network architecture.

\subsection{Experimental Results}
\subsubsection{{Knowledge Amalgamation} from Two Teachers}
To verify the effectiveness of our approach, we evaluate the performance of our learned student model that {amalgamates knowledge} from two teacher models and implements classification task of both teachers. 
The following four methods are compared.
\begin{itemize}
\item{{{\bf Ensemble:} We concatenate the score vectors 
from the two teacher models and classify the input sampling by 
assigning the class of the highest score in the concatenated score vector
as the label of the input.}
}
\item{{{\bf Baseline:} {We learn a student model by applying Hinton's knowledge distillation method~\cite{hinton2015distilling}, which has proven a superior performance to its variants on large number ($\geq$100) of classes. Specifically, we stack the score vectors
from the teachers and use the concatenated vector as the target to train the student.}}
}
\item{ {\bf Layer-wise Learning:} After the feature amalgamation step, we conduct only layer-wise parameter learning to obtain the student network parameters.}
\item{ {\bf Joint Learning:} Our complete model, with parameters first layer-wise learned and then jointly learned.} 
\end{itemize}

The comparative results are shown in Table~\ref{table:two_teachers}. 
On all benchmark datasets, our complete method \emph{Joint Learning}
achieves the highest performance among the four methods,
and demands significantly fewer parameters  
than \emph{Ensemble}. 


We also compare the performance of the learned student model with those of the teachers.
Let \emph{part1} and \emph{part2} denote the categories handled 
by the two teachers models, \emph{teacher1} and \emph{teacher2}, respectively, and let \emph{whole}
denote the complete set of categories.
As shown in Table~\ref{table:animals} and Table~\ref{table:vehicles}, 
our complete student model, \emph{joint learning},
in fact outperforms the teacher models on the corresponding subtasks.
For example, on the Stanford Dogs dataset, the student model achieves
a \emph{part1} accuracy of 61.5\% and a \emph{part2} one of 59.6\%,
while those for \emph{teacher1} and \emph{teacher2}, which specialize
in handling \emph{part1} and \emph{part2}, are 60.8\% and 58.5\% respectively.
These interesting and encouraging results show that our approach is indeed able
to learn the amalgamated knowledge 
from both teachers, and the knowledge learned from one teacher benefits the 
classification task of the other.

{
    \begin{table}
    \begin{center}
    \caption{
       Comparing the results of the teacher models and the learned student models
       on Stanford Dogs and CUB-200-2011 dataset. 
        \emph{Layer} denotes {layer-wise parameter learning} strategy, and
        \emph{Joint} denotes {joint learning} strategy .}
    \vspace{0.2cm}
    \label{table:animals}
    \begin{threeparttable}
    \begin{tabular}{p{1.2cm}<{\centering} |p{0.7cm}<{\centering}p{0.7cm}<{\centering}p{0.7cm}<{\centering}|p{0.7cm}<{\centering}p{0.7cm}<{\centering}p{0.7cm}<{\centering}}
    \hline
    \multicolumn{1}{c|}{} & \multicolumn{3}{c|}{\textbf{Stanford Dogs}}    & \multicolumn{3}{c}{\textbf{CUB-200-2011}} \\
    \textbf{Method} & \textbf{whole} & \textbf{part1} & \textbf{part2} & \textbf{whole} & \textbf{part1} & \textbf{part2} \\ 
    \hline
    Teacher1 & - & 60.8\% & - & - & \textbf{53.8}\% & - \\
    \hline
    Teacher2 & - & - & 58.5\%  & - & - & 49.7\% \\
    \hline
    Layer & 38.4\% & 54.1\% & 54.3\% & 31.8\% & 44.8\% & 41.2\% \\
    \hline
    Joint & \textbf{45.3}\% & \textbf{61.5}\% & \textbf{59.6}\%  & \textbf{42.3}\% & 53.2\% & \textbf{50.3}\% \\
    \hline
    \end{tabular}
    \end{threeparttable}
    \end{center}
    \vspace{-0.5cm}
    \end{table}
}

{
    \begin{table}
    \begin{center}
       \caption{Comparing the results of the teacher models and the learned student models
       on FGVC-Aircraft and Cars dataset. 
        \emph{Layer} denotes {layer-wise parameter learning} strategy, and
        \emph{Joint} denotes {joint learning} strategy .}
    \label{table:vehicles}
    \begin{threeparttable}
    \begin{tabular}{p{1.2cm}<{\centering} |p{0.7cm}<{\centering}p{0.7cm}<{\centering}p{0.7cm}<{\centering}|p{0.7cm}<{\centering}p{0.7cm}<{\centering}p{0.75cm}<{\centering}}
    \hline
    \multicolumn{1}{c|}{} & \multicolumn{3}{c|}{\textbf{FGVC-Aircraft}}    & \multicolumn{3}{c}{\textbf{Cars}} \\
    \textbf{Method} & \textbf{whole} & \textbf{part1} & \textbf{part2} & \textbf{whole} & \textbf{part1} & \textbf{part2} \\ 
    \hline
    Teacher1 & - & 67.6\% & - & - & 52.2\% & - \\
    \hline
    Teacher2 & - & - & 58.8\%  & - & - & 50.1\% \\
    \hline
    Layer & 39.8\% & 59.0\% & 50.8\% & 33.6\% & 47.3\% & 43.4\% \\
    \hline
    Joint & \textbf{49.4}\% & \textbf{67.8}\% & \textbf{59.2}\%  & \textbf{40.6}\% & \textbf{53.0}\% & \textbf{50.4}\% \\
    \hline
    \end{tabular}
    \end{threeparttable}
    \end{center}
    \vspace{-0.5cm}
    \end{table}
}

\subsubsection{Knowledge Amalgamation from Multiple Teachers}
We also test the performance of multi-teacher amalgamation.
We first conduct experiments on amalgamating different 
numbers of teacher models. We split the Stanford Dogs dataset 
comprising 120 classes into four even parts, 
each of which contains 30 classes, and then test the classification 
performances on these parts by  amalgamating knowledge from two, 
three and all four teachers using the DFA model.
We show the results in Table~\ref{table:multiple_teachers}.
Interestingly, the more teachers used for amalgamation,
the higher the classification performance is. For example, 
the performance on \emph{part1} increases from 68.7\% 
to 69.9\% and further to 70.3\% for two, three and four teachers.
This again indicates that the potentially complementary knowledge 
from multiple classification tasks indeed benefits each other.

We then compare the two schemes for multi-teacher amalgamation, 
DFA and IFA. Note that the performances of DFA and IFA differ only when amalgamating from
more than two teachers, and thus we compare their performances using three and four teachers.
We show the results in Table~\ref{table:amalga_module}.
The performances of the two strategies are in general much the same where DFA yields slightly better 
results on one part while IFA on the others.
In our supplementary material, we also provide experimental results of IFA with different amalgamating orders.

{
    \begin{table}
    \begin{center}
    \caption{Classification performances of the student models, whose knowledge is amalgamated from different numbers of teachers using DFA.}
    \label{table:multiple_teachers}
    \vspace{0.2cm}
    \begin{threeparttable}
    \begin{tabular}{p{3.0cm}<{\centering}|p{0.8cm}<{\centering}p{0.8cm}<{\centering}p{0.8cm}<{\centering}p{0.8cm}<{\centering}}
    \hline
    \multicolumn{1}{c|}{} & \multicolumn{4}{c}{\textbf{Stanford Dogs}} \\
    \textbf{Method} & \textbf{part1} & \textbf{part2} & \textbf{part3} & \textbf{part4} \\ 
    \hline
    From 2 teachers  & 68.7\% & 66.1\% & - & - \\
    \hline
    From 3 teachers  & 69.9\% & 67.7\%  & 63.8\% & - \\
    \hline
    From 4 teachers  & \textbf{70.3}\% & \textbf{68.0}\%  & \textbf{65.4}\% & 67.3\% \\
    \hline
    \end{tabular}
    \end{threeparttable}
    \end{center}
    \vspace{-0.4cm}
    \end{table}
}
{
    \begin{table}
    \begin{center}
     \caption{Classification performances of the student models, whose knowledge
     is amalgamated from different numbers of teachers using DFA and IFA.}
    \label{table:amalga_module}
    \vspace{0.2cm}
    \begin{threeparttable}
    \begin{tabular}{p{3.0cm}<{\centering}|p{0.8cm}<{\centering}p{0.8cm}<{\centering}p{0.8cm}<{\centering}p{0.8cm}<{\centering}}
    \hline
    \multicolumn{1}{c|}{} & \multicolumn{4}{c}{\textbf{Stanford Dogs}} \\
    \textbf{Method} & \textbf{part1} & \textbf{part2} & \textbf{part3} & \textbf{part4} \\ 
    \hline
    DFA from 3 teachers & \textbf{69.9}\% & 67.7\% & 63.8\% & - \\
    \hline
    IFA from 3 teachers  & 69.6\% & \textbf{69.1}\%  & \textbf{64.7}\% & - \\
    \hline
    \hline
    DFA from 4 teachers & \textbf{70.3}\% & 68.0\%  & 65.4\% & 67.3\% \\
    \hline
    IFA from 4 teachers & 69.9\% & \textbf{68.8}\%  & \textbf{65.5}\% & \textbf{67.9}\% \\
    \hline
    \end{tabular}
    \end{threeparttable}
    \end{center}
    \vspace{-0.2cm}
    \end{table}
}

\subsection{Ablation Study}
We also conduct ablation studies to validate the proposed method as follows.

\subsubsection{Feature Adaption}
To show the effectiveness of FAM, we compare the classification performances 
of the proposed model with FAM turned on and off.
As  shown in Table~\ref{table:feature_adaption}, 
when FAM is turned on, the accuracies are significantly 
higher than those with FAM turned off on all the datasets.
This indicates that the FAM is indeed able to
transform amalgamated features into a form 
that better adapts to non-parametric layers in the network,
and meanwhile preserves and passes 
the critical information to the next layer.

{

    \begin{table}
    \begin{center}
    \vspace{-0.5cm}
    \caption{
Classification performances of the student model with 
and without FAM. 
For simplicity, ``Dogs'' denotes ``Stanford Dogs'', 
``CUB'' denotes ``CUB-200-2011'' and ``Aircraft'' denotes ``FGVC-Aircraft''.}
    \vspace{0.2cm}
    \label{table:feature_adaption}
    \begin{threeparttable}
    \begin{tabular}{p{2.8cm}<{\centering}|p{0.7cm}<{\centering}|p{0.7cm}<{\centering}|p{1.2cm}<{\centering}|p{0.7cm}<{\centering}}
    \hline
    \multicolumn{1}{c|}{\textbf{Method}} & \multicolumn{1}{c|}{\textbf{Dogs}} & \multicolumn{1}{c|}{\textbf{CUB}} & \multicolumn{1}{c|}{\textbf{Aircraft}} & \multicolumn{1}{c}{\textbf{Cars}}\\
    \hline
    W/O FAM & 25.3\% & 33.9\% & 39.7\% & 30.7\% \\
    \hline
    W/ FAM & \textbf{45.3}\% & \textbf{42.6}\%  & \textbf{49.4}\% & \textbf{40.6}\% \\
    \hline
    \end{tabular}
    \end{threeparttable}
    \end{center}
    \vspace{-0.4cm}
    \end{table}
}

\subsubsection{Layer-wise Parameter Learning}
We also investigate the power of the layer-wise learning strategy,
by comparing the student model with and without the 
layer-wise learning, which correspond to the 
\emph{joint learning} method and the \emph{baseline} 
described in the previous section.
We show their training and testing errors versus the epochs 
in Figure~\ref{fig:training_curve}. 
The test error of \emph{joint learning} with layer-wise learning is significantly lower than that of the \emph{baseline} without layer-wise learning. 
In fact, despite not shown here, 
without layer-wise learning the test error 
at 300 epochs remains to be 60.1\%, as indicated 
by the \emph{baseline} in Table~\ref{table:two_teachers}.
We may thus safely conclude that layer-wise learning indeed 
facilitates the training compared to learning from scratch as done for \emph{baseline}.

{
    \begin{figure}
        \centering
        \includegraphics[width=0.9\linewidth]{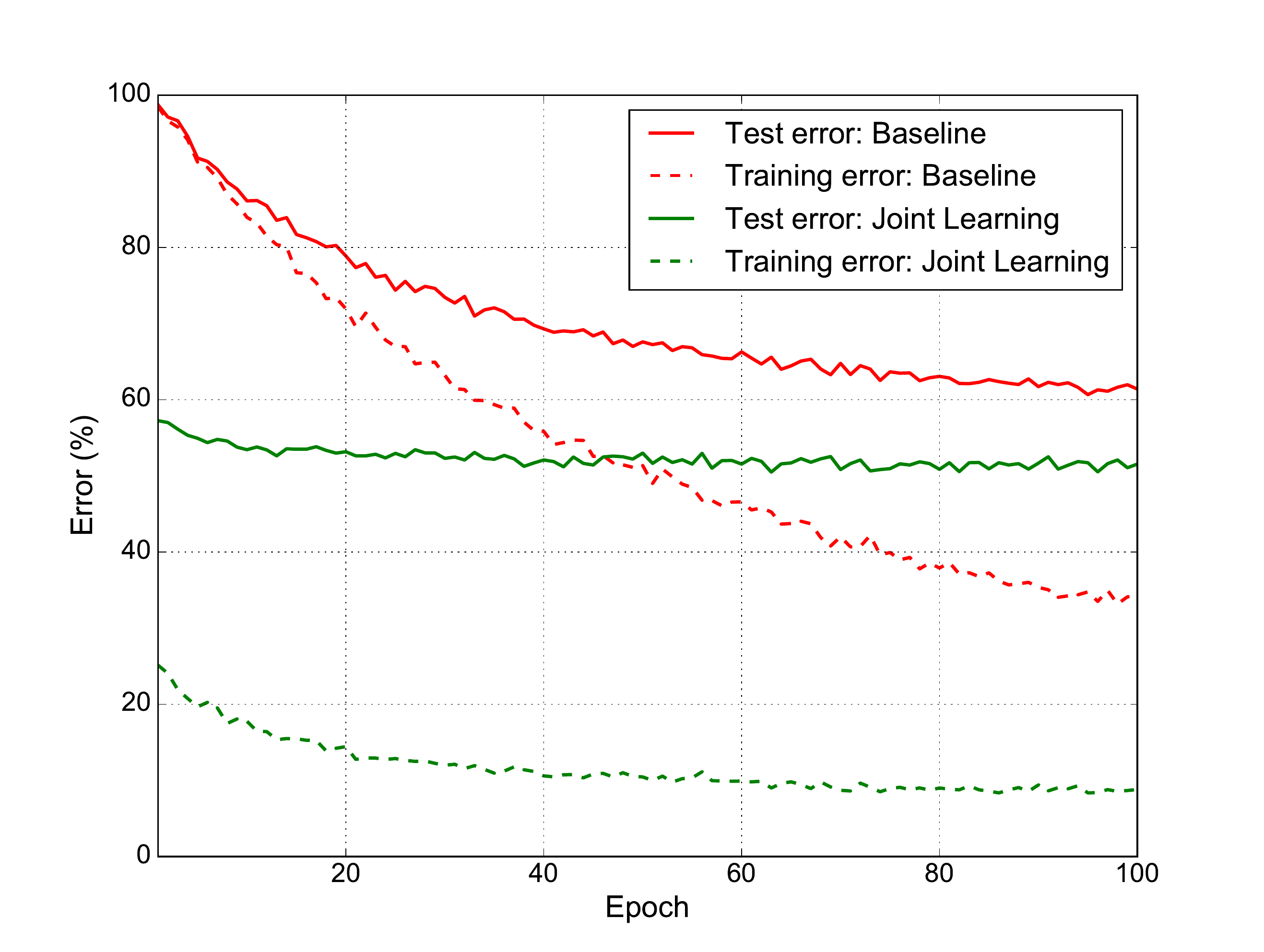}
        \vspace{-0.4cm}
        \caption{
Training and test errors of \emph{baseline} and \emph{joint learning}
versus the number of epochs on Stanford Dogs.
\emph{Joint learning} converges to much lower errors.}
        \label{fig:training_curve}
     \vspace{-0.4cm}
    \end{figure}
}
\subsubsection{Joint Parameter Learning}
We compare in Table~\ref{table:animals} and Table~\ref{table:vehicles} 
the results of \emph{layer-wise learning} only and \emph{joint learning},
where the latter one outperforms the former on all the four datasets.
This validates in part our hypothesis that 
the \emph{layer-wise learning} 
accumulates errors across layers during training,
which can be alleviated by \emph{joint learning}.

\section{Conclusion and Future Work}
In this paper, we propose a new model-reusing task, 
termed \emph{knowledge amalgamation}, 
which aims at learning a compact student model 
capable of handing the ``super'' task from multiple 
teachers, each of which specializes in a different task.
This task is in our opinion a research-worthy topic in 
the sense that it allows amalgamating well-trained models, 
many of which are learned using large-scale or 
private datasets that are not publicly available, 
to derive a lightweight student model
that approximates or even outperforms the teachers.

To this end, we propose a pilot approach towards solving this task.
The proposed approach follows a two-step strategy by first conducting
feature amalgamation from the multiple teachers and then treating
the obtained features as guidance to learn the parameters of the 
student network. We conduct experiments on four datasets to validate 
the proposed approach, which yields very promising results: the 
learned student model can in fact perform better than the teachers 
at their specializations, at a model scale that is much smaller than
the ensemble of the teachers. We also justify the validness of several 
components by conducting the ablation study.

Admittedly, this pilot approach in the current form, despite its very encouraging 
results, indeed has some limitations. We assume that the teacher models and the 
student one share the same network architecture, which might be a strong 
assumption in some real-world scenarios.
In our near-future work, we will therefore investigate amalgamating  knowledge 
from teachers of different network architectures,  which truly allows us to reuse 
the knowledge of massive well-trained neural networks in the wild. 

In the longer term, we will explore how to bridge the semantic gap among different network architectures 
and reuse the amalgamated knowledge to new tasks, enabling the \emph{amalgamated knowledge transfer}.

\subsubsection{Acknowledgments}
This work is supported by  Natonal Basic Research Program of China under Grant No. 2015CB352400, 
National Natural Science Foundation of China (61572428, U1509206), Fundamental Research Funds for 
the Central Universities (2017FZA5014), and Key Research and Development Program of Zhejiang 
Province (2018C01004).

\bibliographystyle{aaai}
\bibliography{paper}

\begin{thebibliography}{}

\bibitem[\protect\citeauthoryear{Buciluǎ, Caruana, and
  Niculescu-Mizil}{2006}]{buciluǎ2006model}
Buciluǎ, C.; Caruana, R.; and Niculescu-Mizil, A.
\newblock 2006.
\newblock Model compression.
\newblock In {\em ACM SIGKDD Conference on Knowledge Discovery and Data Mining
  (KDD)},  535--541.

\bibitem[\protect\citeauthoryear{Cui \bgroup et al\mbox.\egroup
  }{2018}]{cui2018large}
Cui, Y.; Song, Y.; Sun, C.; Howard, A.; and Belongie, S.
\newblock 2018.
\newblock Large scale fine-grained categorization and domain-specific transfer
  learning.
\newblock In {\em The IEEE Conference on Computer Vision and Pattern
  Recognition (CVPR)},  4109--4118.

\bibitem[\protect\citeauthoryear{Ding \bgroup et al\mbox.\egroup
  }{2018}]{ding2018graph}
Ding, Z.; Li, S.; Shao, M.; and Fu, Y.
\newblock 2018.
\newblock Graph adaptive knowledge transfer for unsupervised domain adaptation.
\newblock In {\em European Conference on Computer Vision (ECCV)},  37--52.

\bibitem[\protect\citeauthoryear{Gholami, Rudovic, and
  Pavlovic}{2017}]{gholami2017punda}
Gholami, B.; Rudovic, O.; and Pavlovic, V.
\newblock 2017.
\newblock Punda: Probabilistic unsupervised domain adaptation for knowledge
  transfer across visual categories.
\newblock In {\em The IEEE International Conference on Computer Vision (ICCV)},
   3601--3610.

\bibitem[\protect\citeauthoryear{Gupta, Hoffman, and
  Malik}{2015}]{gupta2015cross}
Gupta, S.; Hoffman, J.; and Malik, J.
\newblock 2015.
\newblock Cross modal distillation for supervision transfer.
\newblock {\em arXiv preprint arXiv:1507.00448}.

\bibitem[\protect\citeauthoryear{He \bgroup et al\mbox.\egroup
  }{2016}]{he2016deep}
He, K.; Zhang, X.; Ren, S.; and Sun, J.
\newblock 2016.
\newblock Deep residual learning for image recognition.
\newblock In {\em The IEEE Conference on Computer Vision and Pattern
  Recognition (CVPR)},  770--778.

\bibitem[\protect\citeauthoryear{Hinton, Vinyals, and
  Dean}{2015}]{hinton2015distilling}
Hinton, G.; Vinyals, O.; and Dean, J.
\newblock 2015.
\newblock Distilling the knowledge in a neural network.
\newblock {\em Advances in Neural Information Processing Systems (NIPS)}.

\bibitem[\protect\citeauthoryear{Hong \bgroup et al\mbox.\egroup
  }{2016}]{hong2016learning}
Hong, S.; Oh, J.; Lee, H.; and Han, B.
\newblock 2016.
\newblock Learning transferrable knowledge for semantic segmentation with deep
  convolutional neural network.
\newblock In {\em The IEEE Conference on Computer Vision and Pattern
  Recognition (CVPR)},  3204--3212.

\bibitem[\protect\citeauthoryear{Hu, Lu, and Tan}{2015}]{hu2015deep}
Hu, J.; Lu, J.; and Tan, Y.-P.
\newblock 2015.
\newblock Deep transfer metric learning.
\newblock In {\em The IEEE Conference on Computer Vision and Pattern
  Recognition (CVPR)},  325--333.

\bibitem[\protect\citeauthoryear{Huang and Wang}{2017}]{NST2017}
Huang, Z., and Wang, N.
\newblock 2017.
\newblock Like what you like: Knowledge distill via neuron selectivity
  transfer.
\newblock {\em arXiv preprint arXiv:1707.01219}.

\bibitem[\protect\citeauthoryear{Huang, Huang, and
  Kr{\"a}henb{\"u}hl}{2018}]{huang2018domain}
Huang, H.; Huang, Q.; and Kr{\"a}henb{\"u}hl, P.
\newblock 2018.
\newblock Domain transfer through deep activation matching.
\newblock In {\em European Conference on Computer Vision (ECCV)},  611--626.

\bibitem[\protect\citeauthoryear{Huang, Peng, and Yuan}{2017}]{ijcai2017-263}
Huang, X.; Peng, Y.; and Yuan, M.
\newblock 2017.
\newblock Cross-modal common representation learning by hybrid transfer
  network.
\newblock In {\em International Joint Conference on Artificial Intelligence
  (IJCAI)},  1893--1900.

\bibitem[\protect\citeauthoryear{Khosla \bgroup et al\mbox.\egroup
  }{2011}]{KhoslaYaoJayadevaprakashFeiFei_FGVC2011}
Khosla, A.; Jayadevaprakash, N.; Yao, B.; and Fei-Fei, L.
\newblock 2011.
\newblock Novel dataset for fine-grained image categorization.
\newblock In {\em First Workshop on Fine-Grained Visual Categorization, IEEE
  Conference on Computer Vision and Pattern Recognition (CVPR)}.

\bibitem[\protect\citeauthoryear{Krause \bgroup et al\mbox.\egroup
  }{2013}]{KrauseStarkDengFei-Fei_3DRR2013}
Krause, J.; Stark, M.; Deng, J.; and Fei-Fei, L.
\newblock 2013.
\newblock 3d object representations for fine-grained categorization.
\newblock In {\em International IEEE Workshop on 3D Representation and
  Recognition (3dRR)}.

\bibitem[\protect\citeauthoryear{Krizhevsky, Sutskever, and
  Hinton}{2012}]{krizhevsky2012imagenet}
Krizhevsky, A.; Sutskever, I.; and Hinton, G.~E.
\newblock 2012.
\newblock Imagenet classification with deep convolutional neural networks.
\newblock In {\em Advances in Neural Information Processing Systems (NIPS)},
  1097--1105.

\bibitem[\protect\citeauthoryear{Long \bgroup et al\mbox.\egroup
  }{2013}]{long2013transfer}
Long, M.; Wang, J.; Ding, G.; Sun, J.; and Philip, S.~Y.
\newblock 2013.
\newblock Transfer feature learning with joint distribution adaptation.
\newblock In {\em The IEEE International Conference on Computer Vision (ICCV)},
   2200--2207.

\bibitem[\protect\citeauthoryear{Maji \bgroup et al\mbox.\egroup
  }{2013}]{maji13fine-grained}
Maji, S.; Kannala, J.; Rahtu, E.; Blaschko, M.; and Vedaldi, A.
\newblock 2013.
\newblock Fine-grained visual classification of aircraft.
\newblock {\em arXiv preprint arXiv:1306.5151}.

\bibitem[\protect\citeauthoryear{Pan, Yang, and others}{2010}]{pan2010survey}
Pan, S.~J.; Yang, Q.; et~al.
\newblock 2010.
\newblock A survey on transfer learning.
\newblock {\em IEEE Transactions on Knowledge and Data Engineering (TKDE)}
  22(10):1345--1359.

\bibitem[\protect\citeauthoryear{Paszke \bgroup et al\mbox.\egroup
  }{2017}]{paszke2017automatic}
Paszke, A.; Gross, S.; Chintala, S.; Chanan, G.; Yang, E.; DeVito, Z.; Lin, Z.;
  Desmaison, A.; Antiga, L.; and Lerer, A.
\newblock 2017.
\newblock Automatic differentiation in pytorch.
\newblock In {\em Advances in Neural Information Processing Systems (NIPS)}.

\bibitem[\protect\citeauthoryear{Romero \bgroup et al\mbox.\egroup
  }{2014}]{romero2015fitnets}
Romero, A.; Ballas, N.; Kahou, S.~E.; Chassang, A.; Gatta, C.; and Bengio, Y.
\newblock 2014.
\newblock Fitnets: Hints for thin deep nets.
\newblock {\em International Conference on Learning Representations (ICLR)}.

\bibitem[\protect\citeauthoryear{Simonyan and
  Zisserman}{2014}]{simonyan2014very}
Simonyan, K., and Zisserman, A.
\newblock 2014.
\newblock Very deep convolutional networks for large-scale image recognition.
\newblock {\em arXiv preprint arXiv:1409.1556}.

\bibitem[\protect\citeauthoryear{Szegedy \bgroup et al\mbox.\egroup
  }{2015}]{szegedy2015going}
Szegedy, C.; Liu, W.; Jia, Y.; Sermanet, P.; Reed, S.; Anguelov, D.; Erhan, D.;
  Vanhoucke, V.; and Rabinovich, A.
\newblock 2015.
\newblock Going deeper with convolutions.
\newblock In {\em The IEEE Conference on Computer Vision and Pattern
  Recognition (CVPR)},  1--9.

\bibitem[\protect\citeauthoryear{Wah \bgroup et al\mbox.\egroup
  }{2011}]{WahCUB_200_2011}
Wah, C.; Branson, S.; Welinder, P.; Perona, P.; and Belongie, S.
\newblock 2011.
\newblock {The Caltech-UCSD Birds-200-2011 Dataset}.
\newblock Technical report.

\bibitem[\protect\citeauthoryear{Wang, Deng, and
  Wang}{2016}]{wang2016accelerating}
Wang, Z.; Deng, Z.; and Wang, S.
\newblock 2016.
\newblock Accelerating convolutional neural networks with dominant
  convolutional kernel and knowledge pre-regression.
\newblock In {\em European Conference on Computer Vision (ECCV)},  533--548.

\bibitem[\protect\citeauthoryear{Xu \bgroup et al\mbox.\egroup
  }{2018}]{xu2018pad}
Xu, D.; Ouyang, W.; Wang, X.; and Sebe, N.
\newblock 2018.
\newblock Pad-net: Multi-tasks guided prediction-and-distillation network for
  simultaneous depth estimation and scene parsing.
\newblock In {\em The IEEE Conference on Computer Vision and Pattern
  Recognition (CVPR)}.

\bibitem[\protect\citeauthoryear{Yang \bgroup et al\mbox.\egroup
  }{2017}]{YangZFJZ17}
Yang, Y.; Zhan, D.; Fan, Y.; Jiang, Y.; and Zhou, Z.
\newblock 2017.
\newblock Deep learning for fixed model reuse.
\newblock In {\em AAAI Conference on Artificial Intelligence (AAAI)},
  2831--2837.

\bibitem[\protect\citeauthoryear{Zagoruyko and
  Komodakis}{2017}]{Zagoruyko2017AT}
Zagoruyko, S., and Komodakis, N.
\newblock 2017.
\newblock Paying more attention to attention: Improving the performance of
  convolutional neural networks via attention transfer.
\newblock In {\em International Conference on Learning Representations (ICLR)}.

\end{thebibliography}

\end{document}